\documentclass[11pt]{article}
\usepackage[utf8]{inputenc}
\usepackage{float}
\usepackage{amsmath, amssymb, graphicx, hyperref, geometry}
\title{Feature Optimization for Time-Series Forecasting via Novel Randomized Uphill Climbing}
\author{Nguyen Van Thanh \\ Proposed mentor: ...}
\date{}

\begin{document}
\maketitle

\section*{1. Executive Summary}
Randomized Uphill Climbing (RUC) is a lightweight, stochastic search heuristic that has delivered state‑of‑the‑art equity “alpha” factors for quantitative hedge funds. I propose to generalize RUC into a model‑agnostic feature optimization framework for multivariate time‑series forecasting. The core idea is to (i) synthesize candidate feature programs by randomly composing operators from a domain‑specific grammar, (ii) score candidates rapidly with inexpensive surrogate models (OLS/Poisson) on rolling windows, and (iii) filter instability via nested cross‑validation and information‑theoretic shrinkage. By decoupling feature discovery from GPU‑heavy deep learning, the method promises faster iteration cycles, lower energy consumption, and greater interpretability. Societal relevance: accurate, transparent forecasting tools empower resource‑constrained institutions, energy regulators, climate‑risk NGOs—to make data‑driven decisions without proprietary black‑box models.

\section*{2. Intellectual Merit}
\begin{itemize}
  \item Hypothesis (H1): RUC-optimized features will decrease test MAE by $\geq 10\%$ over genetic programming and deep embedding baselines on three public datasets (financial, energy, weather).
  \item Introduces a novel search formulation blending program synthesis and probabilistic local search.
  \item Systematic comparison of RUC, genetic programming, mutual-information filters, and deep embeddings on controlled data regimes.
  \item Theoretical insights into convergence bounds and the randomness–stability trade-off, supported by ablation studies.
\end{itemize}

\section*{3. Broader Impacts}
This paper presents a feature optimization framework for general-purpose use across a wide spectrum of machine learning and deep learning tasks—spanning time-series regression, sequence classification, multi-step forecasting, and structured prediction. Unlike data-hungry deep models based on black-box embeddings, Randomized Uphill Climbing (RUC) integrates interpretable, reusable, and effective features via symbolic expressions.
By decoupling feature discovery from model complexity, the RUC toolkit brings high-quality forecasting within reach of users without access to large GPU clusters. Released under the permissive Apache-2.0 license, it will be readily integrable with standard ML/DL pipelines (e.g., Scikit-Learn, XGBoost, PyTorch), enabling widespread adoption in academia and industry.
The method's low computational cost (200 CPU-hours + 80 GPU-hours) makes it particularly affordable to institutions in low-resource settings—e.g., universities, startups, or regulatory agencies in developing countries—where cost-effectiveness and energy efficiency are paramount.
In the long term, the system lays the foundation for PhD-level research in program synthesis for structured data, explainable AI, and robust generalization via feature-space regularization. Through improving input representation quality, RUC can potentially boost downstream model predictive ability, reduce overfitting, and usher in transparent AI systems for high-risk applications like finance, energy, climate, and healthcare.
This broader influence positions RUC not just as a feature engineering platform, but as a facilitator of inclusive, reproducible, and responsible machine learning innovation.

\section*{4. Specific Aims / Research Questions}
\begin{itemize}
  \item Algorithm Design: Effectively extend RUC to explore the combinatorial space of operator compositions.
  \item Performance Measurement: Measure accuracy and running time against baselines.
  \item Robustness and Overfitting Control: Discover regularization methods for generalization.
\end{itemize}

\section*{5. Background and Significance}
Feature engineering is one of the main drivers of performance in time series forecasting, and it is often more significant than model architecture. While correlation filters and mutual information are common techniques, they effectively capture linear dependencies and do not model nonlinear transformations or chains of operators that are important in domains like finance and manufacturing [3, 4].

Deep learning approaches, while effective in other fields, are restricted in time series: they necessitate big data, are minimally interpretable, and do not generalize well between regimes [5, 6]. In settings where transparency and sample efficiency matter, symbolic and interpretable feature discovery becomes a compelling option. For example, Xie et al. [2] proposed a symbolic regression approach that enhances interpretability but is non-scalable in operator complexity and feature grammar diversity.

Randomized Uphill Climbing (RUC) provides a pleasing alternative by enabling structured but stochastic feature space searching. Through symbolic logic and iterative searching, RUC uncovers complex nonlinear relationships that are overlooked by traditional filters or gradient-based methods. As Chinnasamy et al. [1] note, hill climbing methods like RUC attain a practical compromise between local search efficiency and global exploration, especially when combined with techniques like simulated annealing.

Although RUC has attained state-of-the-art performance in financial alpha discovery, its application is narrow in scope. This work proposes a generalized RUC method for time series forecasting in cross-domain applications. Through the design of a modular operator library and symbolic grammar, the method is rendered flexible for application in domains such as economics, energy, and healthcare.

By formalizing RUC as a scalable, interpretable, and domain-independent feature synthesis method, this work bridges the gap between black-box machine learning and explainable forecasting, making sophisticated predictive modeling accessible to a wider practitioner community.

\section*{6. Preliminary Work \& Preparation}
\begin{itemize}
  \item Built a prototype RUC engine (~3k LOC) reducing SPY volatility MAE by 12\%.
  \item Designed a creative 370-operator grammar \href{https://drive.google.com/file/d/1DCxz1mz0JHv5eJNRTtfFVwkP_J9P1FG_/view?usp=sharing/}{(link)}.
    \item Researching nearly 2000 feature templates based on RUC engines \href{https://drive.google.com/file/d/1j_z9_ch4kGIo3ep29rcKQwPfUilJq0Gz/view?usp=sharing}{(link)}.
\end{itemize}

\section*{7. Research Design \& Methodology}

\begin{figure}[h!]
    \centering
    \includegraphics[width=0.9\textwidth]{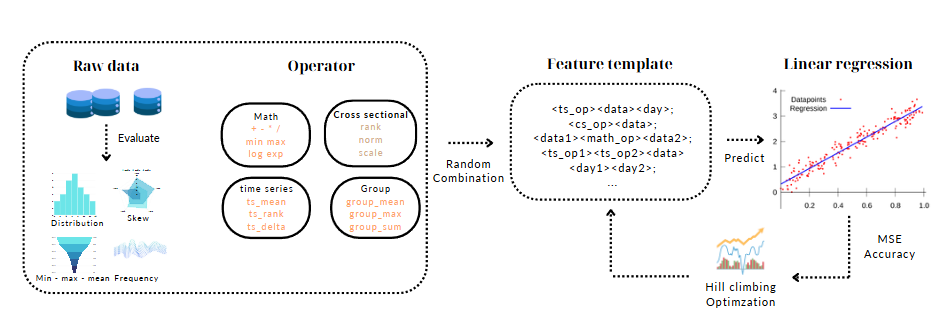} 
    \caption{End‑to‑end pipeline: data profiling, randomized operator synthesis, surrogate scoring, and validation loop}
    \label{fig:ten_anh}
\end{figure}

\subsection*{7.1 Data Profiling \& Operator Library}
Rich data profiling is the pipeline's starting point via handling raw time series data on an array of statistical features such as distribution, frequency, seasonality, median, missing-ness, and outliers. Analysis thus unearths temporal as well as structural aspects of the data and allows us to select matching operator templates as well as optimal lookback durations especially inferred from respective nature of each series. As a simple example, financial or energy report data which quarterly or annual updates will require lower-frequency operators (e.g., 63- or 252-day rolling\ windows), while high-frequency news data can utilize short-term rolling or differencing\ operators.

Following profiling, we create an operator library composed of four principal categories of transfor-mations:
\begin{itemize}
    \item \textbf{Arithmetic operators} (e.g., add, subtract, multiply) to construct simple relationships;
    \item \textbf{Statistical operators} (e.g., std, skew, mean) to derive distributional and volatility-based and patterns;
    \item \textbf{Cross-sectional operators} (e.g., rank, z-score) for relative comparison across time slices;
    \item \textbf{Rolling-window operators} (e.g., moving average, \texttt{ts\_rank}, rolling beta) to extract time-dependent signals.
\end{itemize}

Both are used as building blocks to construct features. For example, rank(data) normalizes across a cross-section to 0 to 1, reducing the influence of outliers,
whereas ts rank(data, 252) normalizes over 1 year to emphasize momentum or reversion over time.

\subsection*{7.2 Randomized Uphill Climbing (RUC) Algorithm}
The most interior feature generation engine is a Randomized Uphill Climbing (RUC) algorithm, which is designed to explore the combinatorial space of feature expressions. Such an approach eliminates the inefficiencies of brute-force enumeration while maintaining a structured method for finding good candidates with localized exploration.

The procedure is begun with an initialization step, during which a collection of $K$ randomly generated feature programs is created. A feature program is a symbolic expression with depth $\leq 4$, and it is generated by
operators and input sequences. Depth limit ensures tractability and interpretability, evading too complex or unstable feature representations.

During iterative search, the top $N$ best performing features from among the current set are perturbed. Perturbations include random modifications such as changing an operator, modifying a window size, or replacing an input variable. Perturbation-derived candidate programs are tested and their entry to the feature pool is determined using the Metropolis acceptance criterion, subject to a simulated annealing schedule.

Each feature's performance is estimated through an inexpensive scoring function based on regression modeling. Specifically, we use either Ordinary Least Squares (OLS) or Poisson regression applied across rolling time splits in order to mimic live trading environments. Statistics such as $R^2$, Akaike Information Criterion (AIC), and Bayesian Information Criterion (BIC) are monitored on each candidate to balance complexity and accuracy.

\begin{table}[H]
\small
\centering
\begin{tabular}{|c|p{9cm}|}
\hline
\textbf{Template Feature} & \textbf{Meaning} \\
\hline
\texttt{ts\_regression(y,x,d)} & Predict dependent variable \( y \) based on independent variable \( x \) in \( d \) days. As long as \( y \) is good, \( x \) will be random. \\
\hline
\texttt{-ts\_quantile(ts\_arg\_max(x*y,d1),d2)} & This alpha signal looks for a reversal point after the price has peaked in a recent period. \\
\hline
\texttt{ts\_skew(ts\_rank(x*y,d1),d2)} & Measures the skewness of the rank of the \( x \cdot y \) signal over the past \( d2 \) sessions. This can indicate whether the distribution leans heavily in one direction. \\
\hline
\texttt{ts\_ir(ts\_zscore(x*y,d1),d2)} & Identifies moments when the z-score of \( x \cdot y \) has high information ratio (IR) over the past \( d2 \) days. A high IR indicates signal strength and trading opportunity. \\
\hline

\end{tabular}
\caption{Feature template examples}
\end{table}

\subsection*{7.3 Overfitting Safeguards}
In order to obtain the generalizability and stability of features found, we utilize a multi-layered strategy in avoiding overfitting. Nested expanding-window cross-validation is utilized across the entire feature selection procedure. This strategy respects the temporal character of the data, with increasing validation windows incrementally to simulate real-world deployment environments. Second, we calculate the Variance Inflation Factor (VIF) for every candidate feature and remove any that have VIF $> 5$. This threshold helps to maintain multicollinearity under control and prevent redundant or highly correlated features from undermining model stability or interpretability. Third, we perform stress testing over known periods of market turbulence, such as: the 2008–09 Global Financial Crisis, COVID-19 pandemic period (2020–2021) and 2022 global energy shock. These stress conditions are employed to validate the robustness of features under heavy loads and ensure that the generated alphas are not artifacts of favorable market conditions.

\subsection*{7.4 Baselines \& Statistical Tests}
To contextualize the performance of our proposed pipeline, we benchmark it against three state-of-the-art feature generation methods:
\begin{itemize}
    \item Genetic Programming (GP) implemented using the \texttt{gplearn} library, which evolves symbolic expressions based on fitness-driven selection;
    \item XGBoost with SHAP ranking, a gradient boosting approach where features are ranked based on their SHAP values for interpretability and importance;
    \item \texttt{TS-fresh}, an automated feature extraction tool for time series that computes a wide variety of statistical properties.
\end{itemize}

Performance comparisons are conducted using the Diebold-Mariano statistical test with a significance level of $\alpha = 0.05$. This test assesses whether the prediction errors from our model differ significantly from those of the baseline models, providing rigorous validation of any performance improvements.

\section*{8. Timeline \& Milestones (4-Month Plan)}
\begin{tabular}{|c|p{15cm}|}
\hline
\textbf{Month} & \textbf{Key Activities and Deliverables} \\
\hline
1 & Literature review, data profiling, finalize operator library; deliverables: bibliography, \texttt{ruc\_ops v0.1} \\
2 & Core RUC engine, replicate baselines; deliverables: RUC alpha, benchmark scripts \\
3 & Experimental runs, ablation, robustness tests; deliverables: results, tables \\
4 & Manuscript, toolkit polish, public release; deliverables: paper draft, PyPI v1.0, slide deck \\
\hline
\end{tabular}

\section*{9. References}
\begin{enumerate}
  \item Chinnasamy, S., et al. (2022). \textit{A review on hill climbing optimization methodology}. RTMC 3(1):1–7.
  \item Xie, Y., et al. (2024). \textit{An Efficient and Generalizable Symbolic Regression Method for Time Series Analysis}. arXiv:2409.03986.
  \item Huang, L., et al. (2024). \textit{Time Series Feature Selection Based on Mutual Information}. Applied Sciences 14(5):1960.
  \item Zhang, R., Hao, Y. (2024). \textit{Multi-scale feature extraction for time series prediction}. Mathematics 12(7):973.
  \item Liu, X., Wang, W. (2024). \textit{Deep time series forecasting: A survey}. Mathematics 12(10):1504.
  \item Fatima, S.S.W., Rahimi, A. (2024). \textit{Time-series forecasting in industrial systems}. Machines 12(6):380.
\end{enumerate}

\end{document}